\documentclass{article}
\usepackage{spconf,amsmath,graphicx}

\usepackage{enumitem}
\setlist{nosep, leftmargin=*}
\usepackage[table]{xcolor}
\usepackage{amssymb, amsfonts}
\usepackage{euscript}
\usepackage{xcolor}
\usepackage{hyperref}
\usepackage{cleveref}
\usepackage{mathtools}
\usepackage{booktabs}
\usepackage{multirow}
\usepackage{threeparttable}
\usepackage{subcaption}
\usepackage{hyperref}
\graphicspath{Figures/}
\usepackage{graphicx}
\usepackage[export]{adjustbox}

\title{Leveraging Unlabeled Data for 3D Medical Image Segmentation through Self-Supervised Contrastive Learning}

 \name{Sanaz Karimijafarbigloo$^{1}$ \qquad Reza Azad$^{2}$ \qquad Yury Velichko$^{3}$ \qquad Ulas Bagci$^{3}$ \qquad Dorit Merhof$^1$}

 \address{$^{1}$ Faculty of Informatics and Data Science, University of Regensburg, Germany
 \\ $^{2}$ Faculty of Electrical Engineering and Information Technology, RWTH Aachen University, Germany
 \\ $^{3}$Machine and Hybrid Intelligence Lab, Northwestern University, Chicago, IL, USA}

\begin{document}
%
\maketitle
\begin{abstract}
Current 3D semi-supervised segmentation methods face significant challenges such as limited consideration of contextual information and the inability to generate reliable pseudo-labels for effective unsupervised data use. To address these challenges, we introduce two distinct subnetworks designed to explore and exploit the discrepancies between them, ultimately correcting the erroneous prediction results. More specifically, we identify regions of inconsistent predictions and initiate a targeted verification training process. This procedure strategically fine-tunes and harmonizes the predictions of the subnetworks, leading to enhanced utilization of contextual information. Furthermore, to adaptively fine-tune the network's representational capacity and reduce prediction uncertainty, we employ a self-supervised contrastive learning paradigm. For this, we use the network's confidence to distinguish between reliable and unreliable predictions. The model is then trained to effectively minimize unreliable predictions. Our experimental results for organ segmentation, obtained from clinical MRI and CT scans, demonstrate the effectiveness of our approach when compared to state-of-the-art methods. The codebase is accessible on \href{https://github.com/xmindflow/SSL-contrastive}{GitHub}.
\end{abstract}
\section{Introduction}
\label{sec:intro}
Image segmentation, the process of partitioning an image into distinct regions, is of great importance in various medical applications. Consequently, it serves as a fundamental tool for automating clinical workflows, reducing data processing time, and providing quantitative measures of organs or pathologies~\cite{azad2022medical,azad2023foundational}. These capabilities greatly assist clinicians in making more accurate diagnoses, assessing the response to therapy, and ultimately improving patient care and outcomes. In this context, semantic segmentation is essential for improving the quality of medical image segmentation, identifying relevant regions of interest (ROIs), such as tumors or organs~\cite{antonelli2022medical,srivastava2022efficient}, and removing unwanted objects. 
However, conventional supervised approaches depend on obtaining large-scale annotated data, which can be immensely costly in real-world medical scenarios. To address this challenge, a promising solution is semi-supervised semantic segmentation, where models are trained using a limited number of labeled samples and an abundance of unlabeled data. Effectively leveraging unlabeled data is a crucial in this context, driving researchers to explore innovative approaches~\cite{cps,tarvainen2017mean,wang2023mcf,cct}.

One commonly used approach to address this challenge is the application of pseudo-labeling~\cite{lee2013pseudo}. Pseudo-labels are assigned to unlabeled pixels based on predictions from a model trained on labeled data. These "pseudo-labels" then guide the training of a supervised model, improving its performance.
In this regard, Basak et al.~\cite{basak2023pseudo} proposed a method for semantic segmentation that combines contrastive learning (CL) and semi-supervised learning (SemiSL) without requiring a specific initial task. Their method employs pseudo-labels from SemiSL to enhance CL guidance, leading to more accurate multi-class segmentation by learning discriminative class information. Chai et al.~\cite{chaitanya2023local} proposed a local contrastive loss approach to improve pixel-level feature learning for segmentation tasks. Their method leverages semantic label information obtained from pseudo-labels of unlabeled images, in conjunction with a limited set of annotated images with ground truth labels. 
In another study, Bai et al.~\cite{bai2023bidirectional} addresses empirical mismatch issues in semi-supervised medical image segmentation by bidirectionally copying and pasting labeled and unlabeled data in a mean teacher architecture. This promoted consistent learning between labeled and unlabeled data, effectively reducing the empirical distribution gap for improved segmentation performance.
Wang et. al.~\cite{wang2023mcf} introduced a two-stream network as a to address errors within each subnet. This innovative approach considers the discrepancies between pseudo-labels and predictions, effectively rectifying mistakes made by individual networks. 
Similarly, Luo et al.~\cite{luo2021semi} and Wu et al.~\cite{wu2021semi} adopted a dual-task consistency and mutual consistency training strategy to penalize incorrect predictions made by the networks.

Despite the effectiveness of the pseudo-labeling paradigm, concerns remain about the reliability of pseudo labels, which can lead to inaccurate mask predictions. Previous research has attempted to mitigate this issue by filtering out predictions with classification scores below a certain threshold~\cite{zhang2021flexmatch}, but this approach may not be entirely effective in eliminating incorrect predictions. Specifically, some wrong predictions can still exhibit high classification scores, resulting in overconfidence or miscalibration phenomena~\cite{guo2017calibration}. 
Additionally, setting a high threshold to remove unreliable predictions can significantly reduce the number of generated pseudo-labels, limiting the effectiveness of the semi-supervised learning algorithm. This reduction in pseudo-labels can lead to categorically imbalanced training data, which can cause problems like inaccurate assignment of pseudo-labels to pixels related to a particular organ or tissue type. As a result, this imbalance can negatively impact the overall segmentation performance.

As discussed before, directly using unreliable predictions as pseudo-labels can adversely affect the model performance. To address this challenge, we thus introduce here an innovative alternative approach that effectively leverages unreliable pseudo-labels while overcoming the limitations of their direct application. To this end, we propose a dual-stream network architecture in which each subnetwork employs a 3D auto-encoder-decoder module to generate segmentation maps for input images. A supervised loss function guides the network in learning representations for each class, enabling precise and dense predictions. We also propose a consistency regularization term to penalize inaccurate predictions made by each network, using a set of confidence predictions obtained from both pathways. This strategic approach allows the model to adaptively update its feature representations, reducing the occurrence of incorrect predictions from both paths. 
To effectively utilize unlabeled data, we extend the concept of pseudo-labeling by distinguishing between reliable and unreliable predictions. We then optimize the clustering space using a contrastive method to align feature descriptions of unreliable pixels with positive prototypes derived from trustworthy predictions.

Our contributions can be summarized as follows: 
\textbf{(1)} we introduce a consistency regularization term to reduce false predictions;
\textbf{(2)} we conceptualize a self-supervised contrastive learning paradigm to decrease the number of unreliable predictions;
\textbf{(3)}: we obtained the state-of-the-art (SOTA) results on 3D CT/MRI segmentation datasets.

\section{Proposed Method}
\subsection{Overview}
Our goal is to train a semantic segmentation model using a combination of labeled ($\mathcal{D}_l=\{(\mathbf{x}_i^l, \mathbf{y}_i^l)\}_{i=1}^{N_l}$) and larger unlabeled data ($\mathcal{D}_u=\{\mathbf{x}_i^u\}_{i=1}^{N_u}$). To achieve this, we employ two subnetworks, Subnet A ($f_A(x;\theta_1)$) and Subnet B ($f_B(x;\theta_2)$), each employing a 3D encoder-decoder architecture. These subnetworks generate prediction maps (denoted as $Y_A$ and $Y_B$) and corresponding feature representations for each voxel ($v$) in a $D$-dimensional space.

\begin{figure*}[t]
    \centering
    \includegraphics[width=0.9\textwidth]{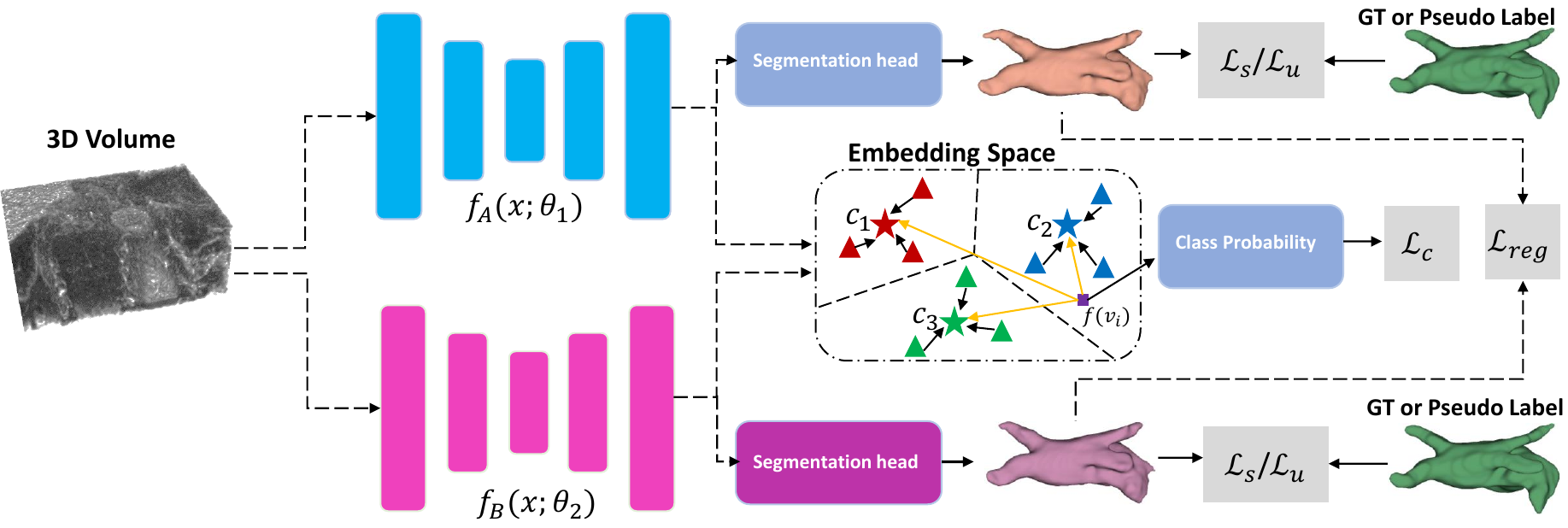}
    \caption{An illustration of our suggested pipeline. In each itteration, we utilize $\mathcal{L}_s$ for the labeled data and $\mathcal{L}_u$ for the unlabeled data. When dealing with the unlabeled data, we adopt the prediction of the network with the lower $\mathcal{L}_s$ as a pseudo label.}
    \label{fig:overview}
    \vspace{-0.8em}
\end{figure*}

During each training step, we randomly sample $b$ labeled images ($b_l$) and $b$ unlabeled images ($b_u$).
For the labeled images, our primary objective is to minimize the standard cross-entropy loss and Dice loss as defined in \Cref{eq:suploss}:

\begin{equation}
\label{eq:suploss}
\begin{aligned}
\mathcal{L}_s = \frac{1}{|\mathcal{B}_l|} \sum_{(\mathbf{x}_i^l, \mathbf{y}_i^l) \in \mathcal{B}_l} \ell_{ce}(f(\mathbf{x}_i^l; \theta), \mathbf{y}_i^l)
+ \text{Dice}({\hat{\mathbf{y_i}}^l}, \mathbf{y_i}^l),
\end{aligned}
\end{equation}

\noindent where $\mathbf{y}_i^l$ represents the hand-annotated mask label for the $i$-th labeled image. Additionally, to minimize false prediction, we introduce consistency regularization term that considers the confidence predictions of one network against the other. We further employ the contrastive loss function to effectively leverage unreliable predictions through the training process. \Cref{fig:overview} illustrates the overall network process.

For the unlabeled images, we feed them through both networks to obtain prediction maps, $Y_a$ and $Y_b$. We then apply a pixel-level entropy-based filtering to exclude unreliable pixel-level pseudo-labels when calculating the unsupervised loss defined in Equation \ref{eq:unsloss}: 

\begin{equation}
\label{eq:unsloss}
\mathcal{L}_u = \frac{1}{|\mathcal{B}_u|} \sum_{\mathbf{x}_i^u \in \mathcal{B}_u} \ell_{ce}(f (\mathbf{x}_i^u; \theta), \hat{\mathbf{y}}_i^u)
+ \text{Dice}(\hat{\mathbf{y}_i}, \mathbf{\hat{\mathbf{y}}_i^u})+\mathcal{L}_{reg},
\end{equation}

\noindent where $\hat{\mathbf{y}}_i^u$ is the pseudo-label for the $i$-th unlabeled image. To further perform error correction, we introduce a regularization loss, $\mathcal{L}_{reg}$, to the $\mathcal{L}_u$, as detailed in \Cref{sec:pseudo}.
Finally, we employ a contrastive loss to exploit unreliable pixels excluded from the unsupervised loss, as explained in Section \Cref{sec:contrastive}.
Our optimization objective is to minimize the overall loss as follows:

\begin{equation}
\label{eq:combo}
\mathcal{L} = \mathcal{L}_s + \lambda_u \mathcal{L}_u + \lambda_c \mathcal{L}_c,
\end{equation}

Here, $\mathcal{L}_s$ and $\mathcal{L}_u$ represent the supervised and unsupervised losses applied to labeled and unlabeled images, respectively, while $\mathcal{L}_c$ is the contrastive loss for unreliable pseudo-labels. The weights $\lambda_u$ and $\lambda_c$ control the contributions of the unsupervised loss and contrastive loss, respectively.
Both $\mathcal{L}_s$ and $\mathcal{L}_u$ are computed using the combination of cross-entropy (CE) and Dice losses  as shown in Figure \Cref{fig:overview}.

\subsection{Consistency Regularization}
\label{sec:pseudo}
We introduce a consistency regularization mechanism to refine the predictions of our dual-subnet model. This module identifies areas where the subnetworks (Subnet A and Subnet B) make discrepant predictions despite being highly confident in their respective predictions, indicating potential mispredictions. The goal is to rectify these incongruities.
Mathematically, we define the area of incorrect predictions between the softmax outputs $\hat{Y}^A$ and $\hat{Y}^B$ as follows:

\begin{equation}
M_{\text{diff}} = \arg_{\max}(\max(\hat{Y}^u_A) > T) \neq \arg_{\max}(\max(\hat{Y}^u_B) > T),
\end{equation}

\noindent where $M_{\text{diff}}$ represents the set of voxels in which Subnet A and Subnet B generate different predictions with high confidence and T represents the confidence threshold. We dynamically adjust the value of T during the training process. We then define the $L1$ distance loss function as regularization term to correct potential incorrect predictions by each of the networks:
\begin{equation}
\mathcal{L}_{reg}  = \sum_{i=1}^{n} |(M_{\text{diff}} \odot \hat{Y}^u_A) - (M_{\text{diff}} \odot \hat{Y}^u_B)|,
\end{equation}
where $\odot$ denotes the Hadamard multiplication.

\subsection{Contrastive loss}
\label{sec:contrastive}
\begin{figure}[h]
    \centering
    \resizebox{\linewidth}{!}{
    \includegraphics[width=\linewidth]{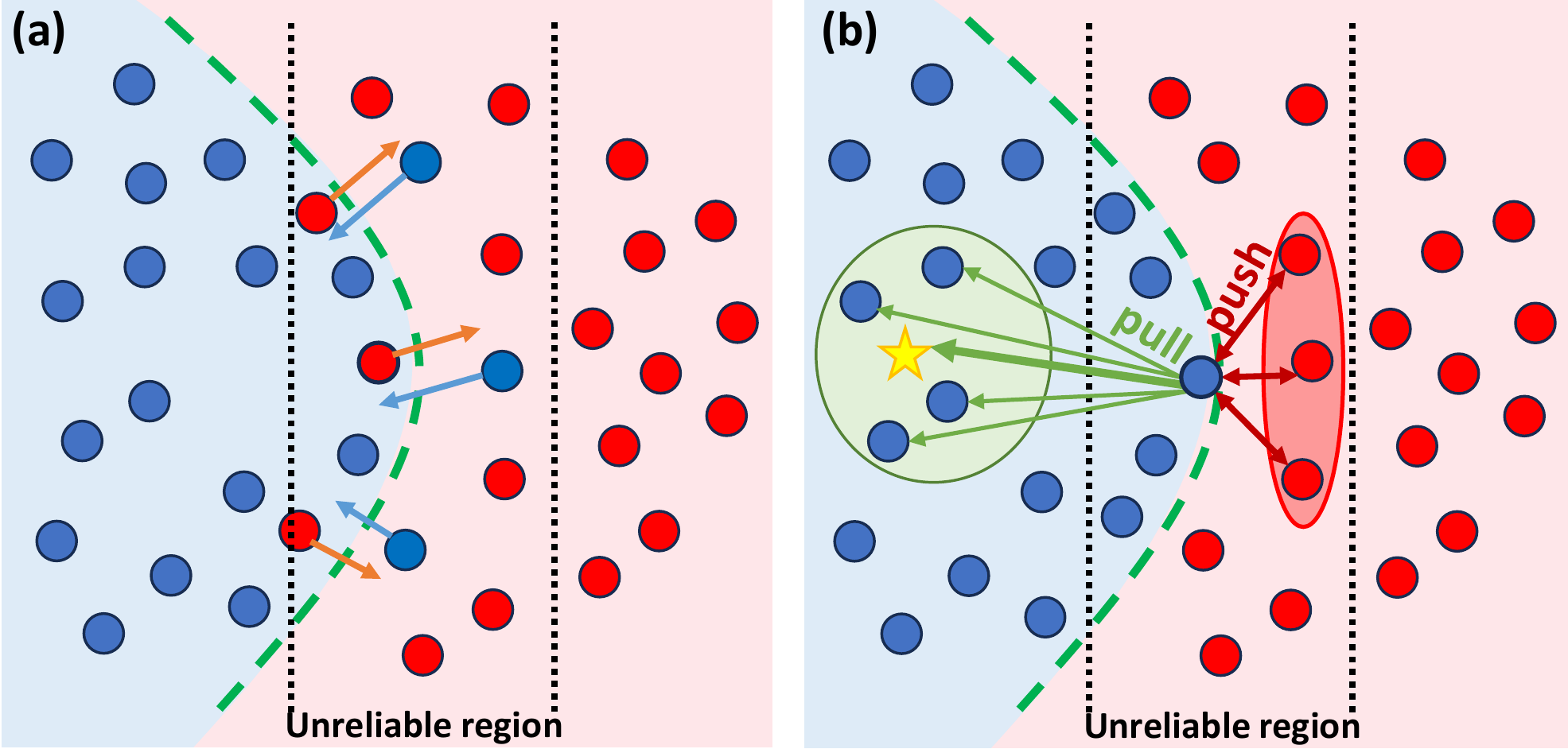}
    }
    \caption{(a): Illustration of the regularization term and (b) contrastive loss effects on prediction refinement.}
    \label{fig:fig2}
    \vspace{-1em}
\end{figure}

To mitigate uncertain predictions in our model, we incorporated the contrastive loss function. \Cref{fig:fig2}b illustrates a scenario where the network's predictions exhibit low confidence in categorizing certain voxels. Our contrastive loss design aims to guide these uncertain voxels towards aligning with their corresponding class prototypes, ultimately reducing misclassifications and uncertainty rates.
To achieve this, our approach first computes the confidence of each voxel's prediction. It then categorises the predictions into two distinct sets: reliable and unreliable predictions. Next, it defines prototypes for each category using the reliable set as a base. Each prototype is computed as the mean vector of the reliable voxel representations:
\begin{equation}
\mathbf{c}_k = \frac{1}{\left|S_k\right|} \sum_{\left(\mathbf{v^r}_i, y_i\right) \in S_k} f(\mathbf{v^r}_i),
\end{equation}
where, $f(\mathbf{v^r}_i)$ indicates the feature represenation of the voxel corresponding to the reliable predictions.
Our approach uses a distance function, denoted as \(d: \mathbb{R}^M \times \mathbb{R}^M \rightarrow [0,+\infty)\), to compute a distribution over classes for uncertain voxels \(\mathbf{v^u}\). This distribution is computed by applying a softmax operation to the distances between the voxel's representation in the embedding space and the class prototypes:
\begin{equation}
p_{\boldsymbol{\phi}}(y=k \mid \mathbf{v^u}) = \frac{\exp \left(-d\left(f(\mathbf{v^u}), \mathbf{c}_k\right)\right)}{\sum_{k^{\prime}} \exp \left(-d\left(f(\mathbf{v^u}), \mathbf{c}_{k^{\prime}}\right)\right)}
\end{equation}

Our contrastive loss function aims to move uncertain voxels of the same class towards their respective class prototype, while also pushing the prototypes of each class away from each other to account for the distance between them. 

\section{Experimental Results}
\subsection{Experimental Setup}
In our study, we developed and implemented our model using the PyTorch framework on a single RTX 3090 GPU. Following \cite{wu2021semi,wang2023mcf} we choose Resnet and V-Net for the two stream network for fair comparision. To optimize the parameters of our network, we employed the SGD optimizer with a weight decay factor of 0.0001 and a momentum coefficient of 0.9. We set our initial learning rate to 0.01 and implemented a dynamic learning rate schedule that reduced the learning rate by a factor of 10 after every 2500 iterations, for a total of 6000 iterations.
During our training process, we included both labeled and unlabeled samples in each iteration, maintaining a consistent batch size of two for both categories.
In \Cref{eq:combo} we use $\lambda_u = 1.0$ and $\lambda_c = 0.1*e^{4(1-t/t_{max})^2}$, where $t$ and $t_{max}$ denote the current and maximum itterations, respectively.
Additionally, for a robust assessment of our model's performance, we adopted the K-fold cross-validation method recommended by \cite{wang2023mcf}.

\subsection{Dataset}
\noindent\textbf{Left Atrial Dataset (LA)}: The LA dataset~\cite{xiong2021global} consists of 1003 3D gadolinium-enhanced MR imaging volumes with manual left atrial annotations, featuring an anisotropic resolution of 0.625 × 0.625 × 0.625 mm³. We preprocessed the data according to~\cite{wang2023mcf}, initially applying volume normalization to standardize the data. During training, we employed random cropping to achieve model input dimensions of 112 × 112 × 80. For inference, we used a sliding window approach with the same dimensions and a stride of 18 × 18 × 4.

\noindent\textbf{NIH Pancreas Dataset}: The NIH Pancreas Dataset~\cite{roth2015deeporgan} comprises 82 abdominal CT volumes with manual pancreas annotations. The CT volumes have dimensions of 512×512×D, where D represents the number of CT slices, which  ranges from 181 to 466. Our preprocessing method, similar to \cite{luo2021semi,wang2023mcf}, involves applying a soft tissue CT window with Hounsfield Units (HU) from -120 to 240. We then align the CT scans to the pancreas region and expand the margins by 25 voxels. During training, we perform random cropping, resulting in volumes with dimensions of 96×96×96. For inference, we use a stride of 16×16×16.

\subsection{Results}

\begin{table}[!htb]
    \centering
    \caption{Comparison of results using the LA dataset (MRI).}
    \label{table1}
    \begin{threeparttable}
    \resizebox{\columnwidth}{!}{
    \begin{tabular}{c|cccc}
        \hline 
        \hline
        Method & Dice(\%)$\uparrow$ & Jaccard(\%)$\uparrow$ & 95HD(voxel)$\downarrow$ & ASD(voxel)$\downarrow$ \\
        \hline
        MT \cite{tarvainen2017mean} & 85.89 $\pm$ 0.024 & 76.58 $\pm$ 0.027 & 12.63 $\pm$ 5.741 & 3.44 $\pm$ 1.382 \\
        UA-MT \cite{yu2019uncertainty} & 85.98 $\pm$ 0.014 & 76.65 $\pm$ 0.017 & 9.86 $\pm$ 2.707 & 2.68 $\pm$ 0.776 \\
        SASSNet \cite{li2020shape} & 86.21 $\pm$ 0.023 & 77.15 $\pm$ 0.024 & 9.80 $\pm$ 1.842 & 2.68 $\pm$ 0.416 \\
        DTC \cite{luo2021semi} & 86.36 $\pm$ 0.023 & 77.25 $\pm$ 0.020 & 9.02 $\pm$ 1.015 & 2.40 $\pm$ 0.223 \\
        MC-Net \cite{wu2021semi} & 87.65 $\pm$ 0.011 & 78.63 $\pm$ 0.013 & 9.70 $\pm$ 2.361 & 3.01 $\pm$ 0.700 \\
        MCF \cite{wang2023mcf}  & 88.71 $\pm$ 0.018 & 80.41 $\pm$ 0.022 & 6.32 $\pm$ 0.800 & 1.90 $\pm$ 0.187 \\
        \rowcolor[HTML]{C8FFFD}
        Our Method & 89.10 $\pm$ 0.012 & 81.62 $\pm$ 0.024 & 6.30 $\pm$ 0.850 & 1.80 $\pm$ 0.020 \\
        \hline
        \hline
    \end{tabular}}
    \end{threeparttable}
\end{table}

\begin{figure}[!thb]
\centering
\resizebox{0.5\textwidth}{!}{
    \begin{tabular}{@{} *{4}c @{}}
    \includegraphics[width=0.25\textwidth, trim=219.5pt 140pt 219.5pt 110pt, clip]{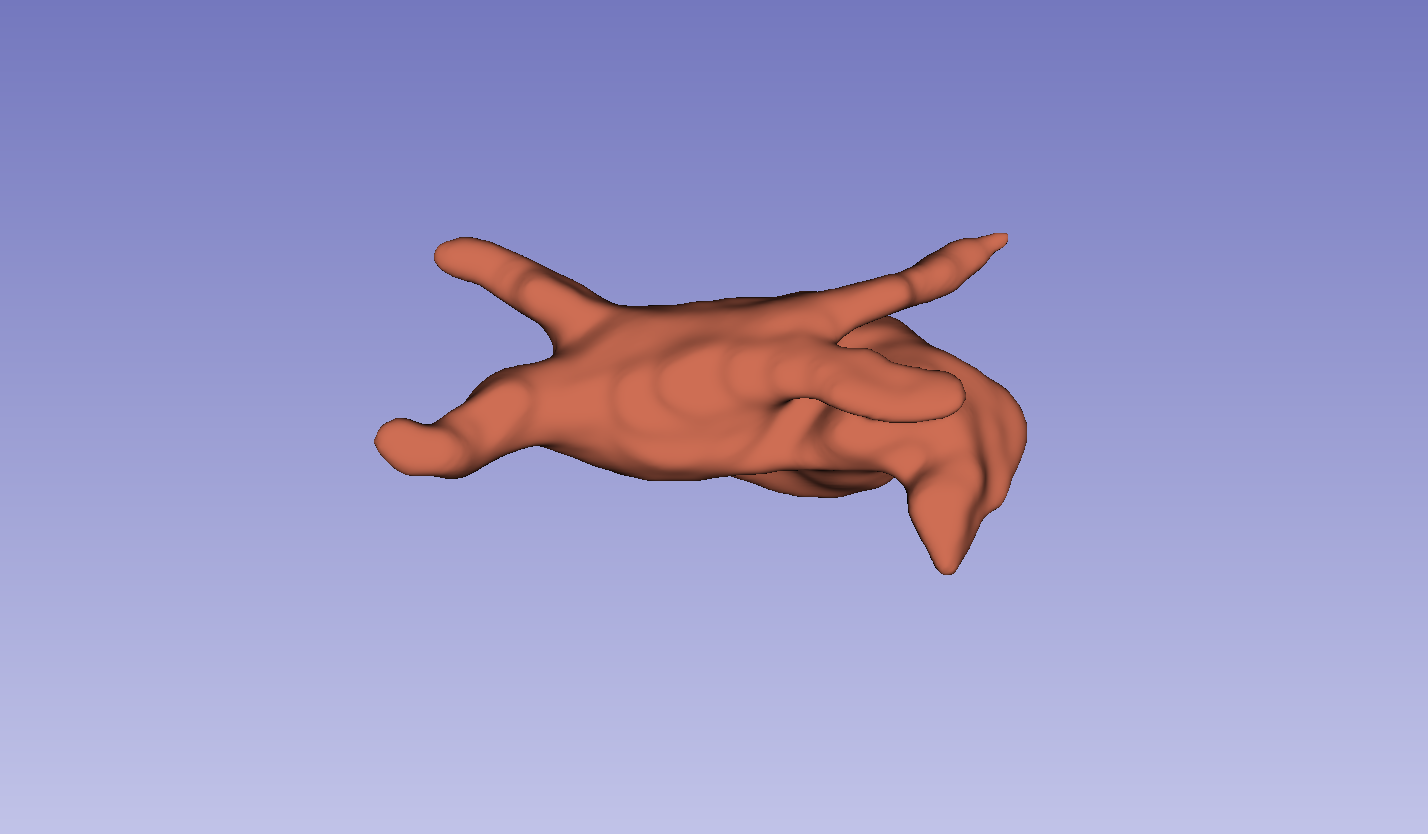} &
    \hspace{-1em} 
    \includegraphics[width=0.25\textwidth, trim=219.5pt 140pt 219.5pt 110pt, clip]{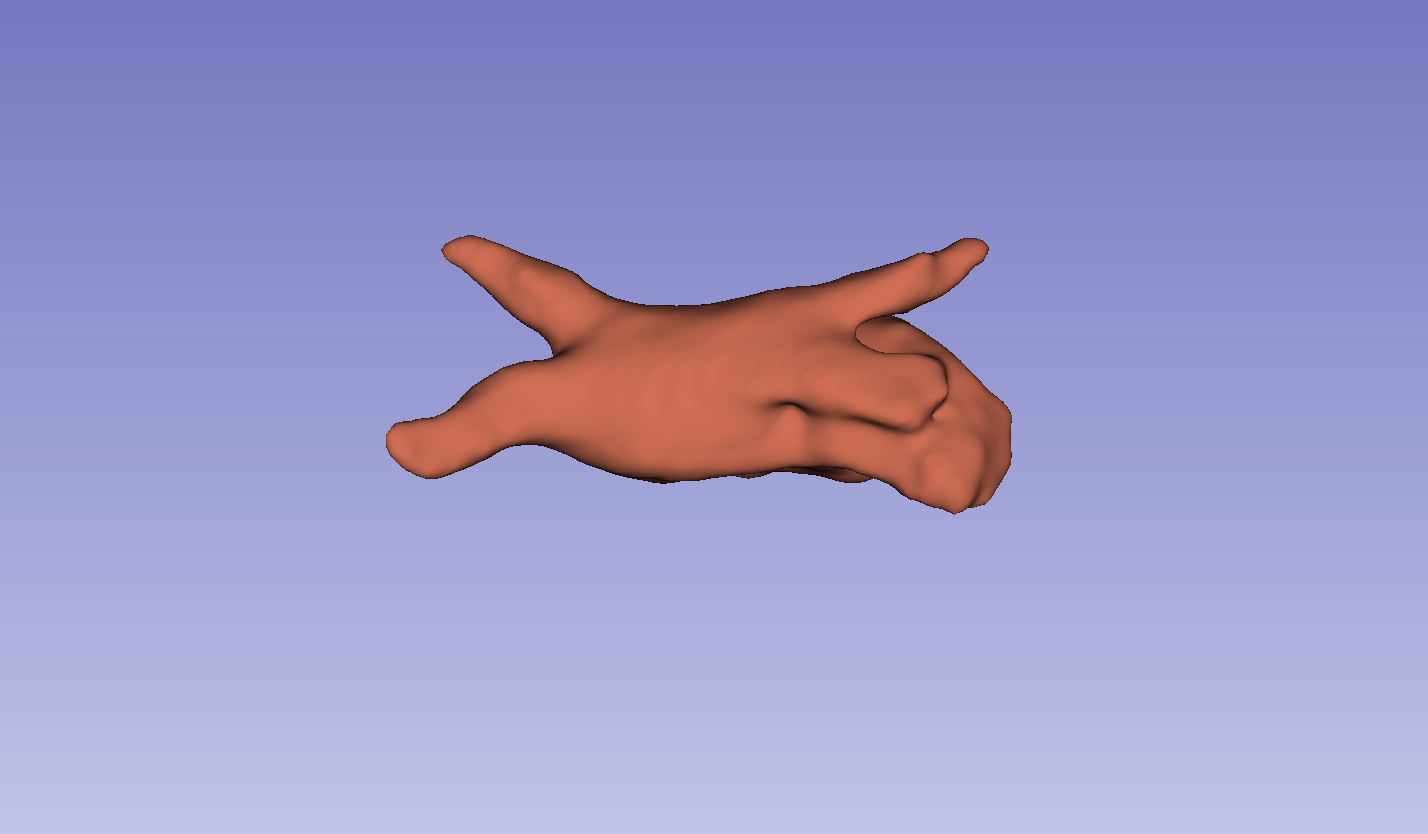} &
    \hspace{-1em} 
    \includegraphics[width=0.25\textwidth, trim=219.5pt 140pt 219.5pt 110pt, clip]{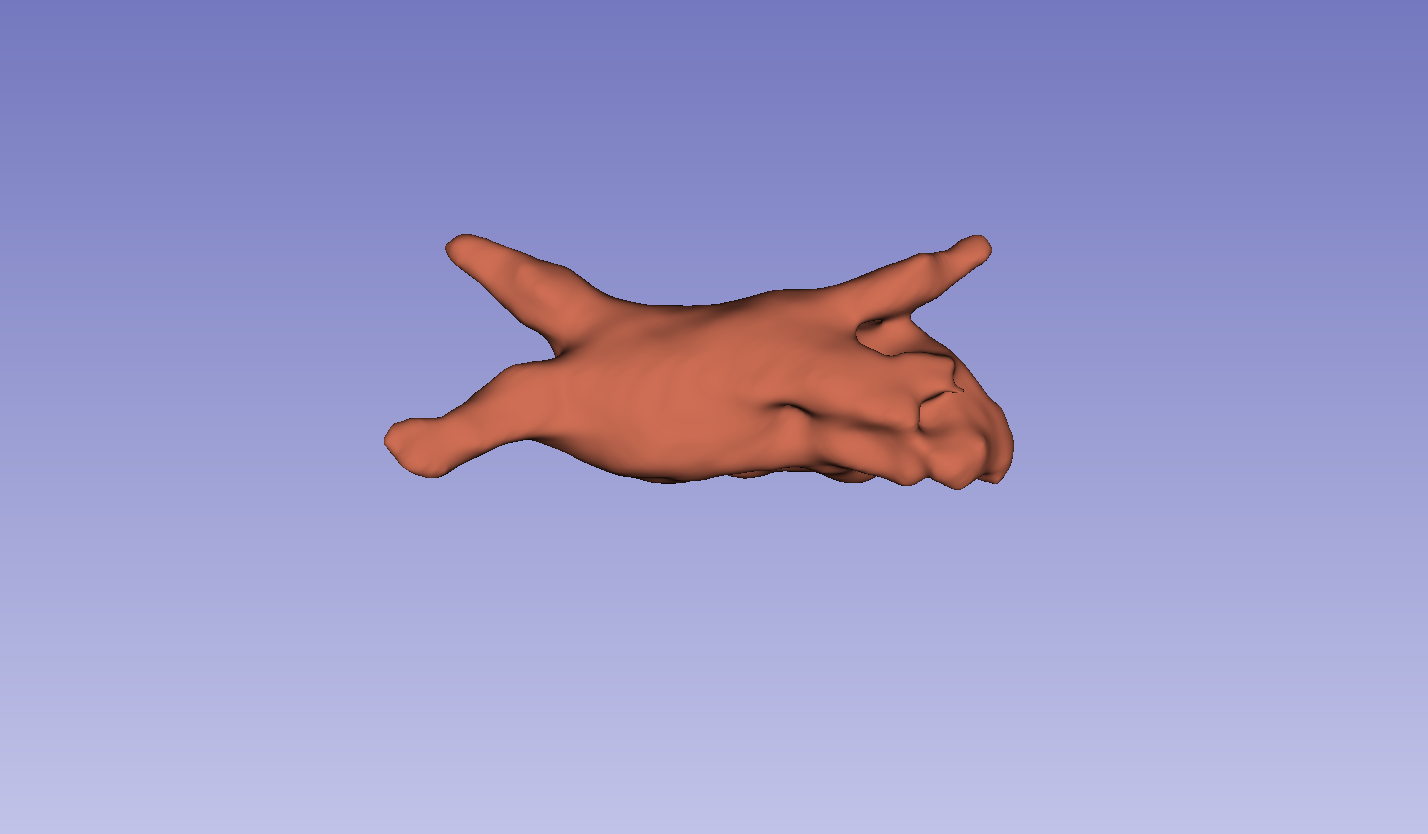} & 
    \hspace{-1em} 
    \includegraphics[width=0.25\textwidth, trim=219.5pt 140pt 219.5pt 110pt, clip]{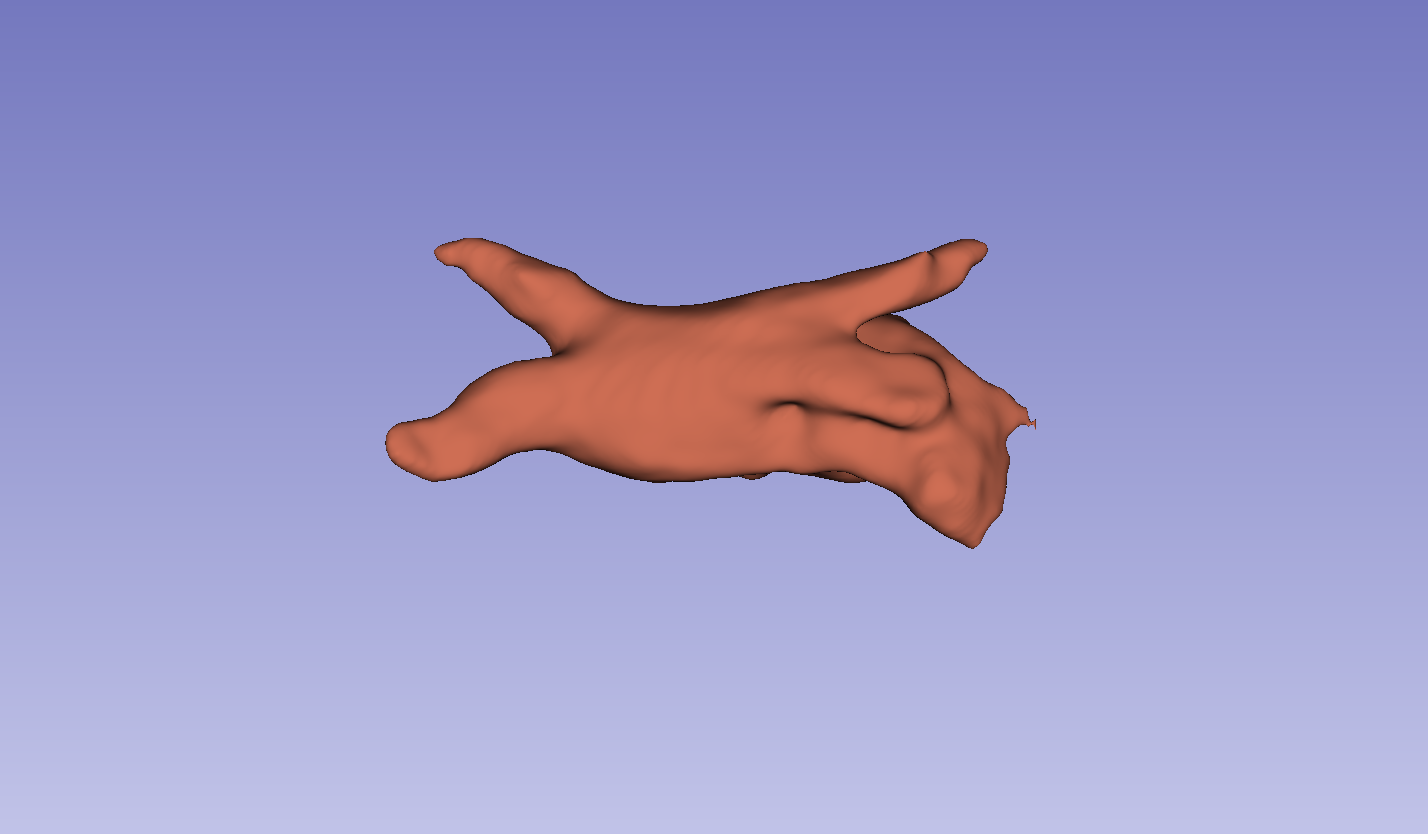} \\
    \includegraphics[width=0.25\textwidth, trim=219.5pt 140pt 219.5pt 110pt, clip]{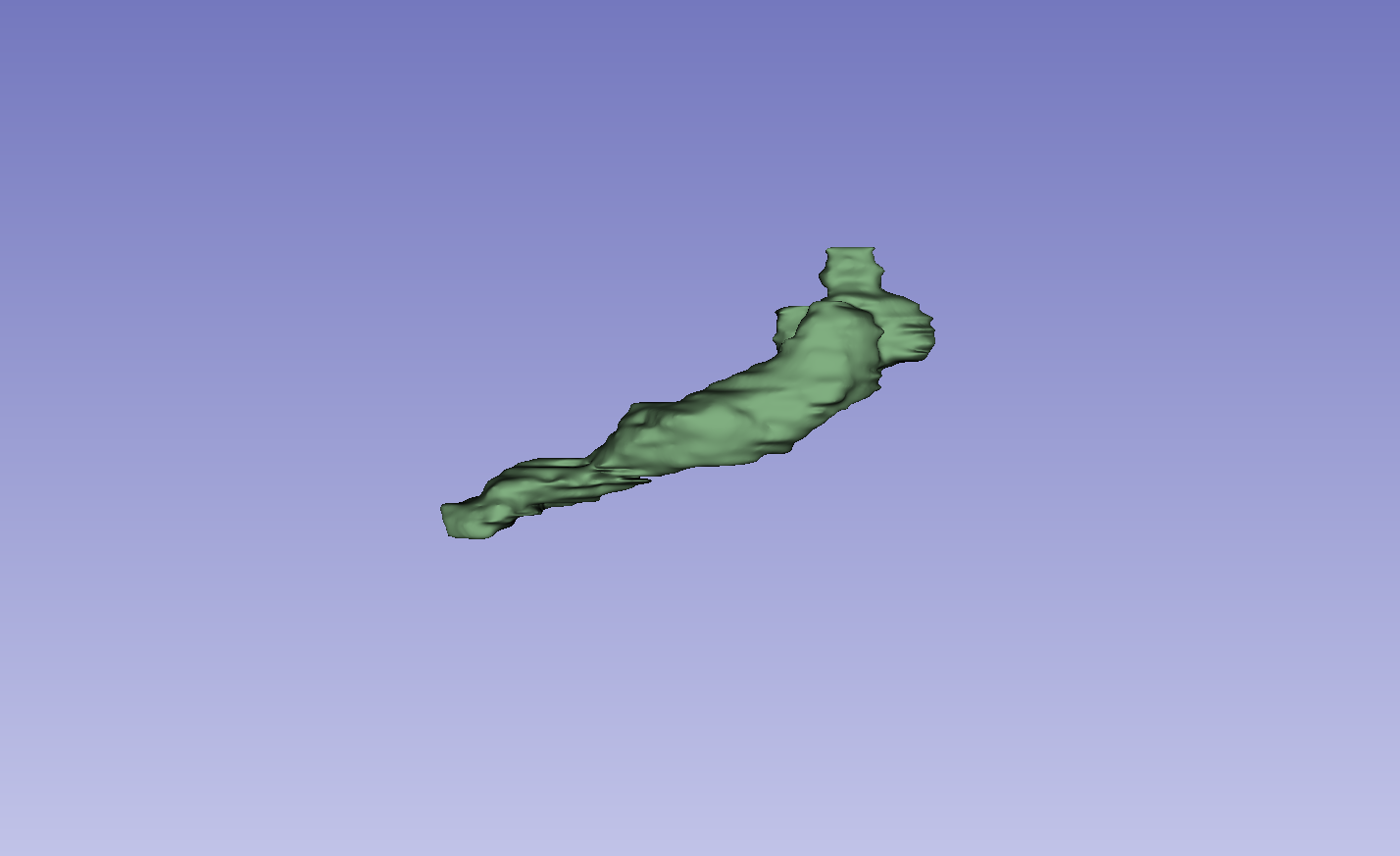} &
    \hspace{-1em} 
    \includegraphics[width=0.25\textwidth, trim=219.5pt 140pt 219.5pt 110pt, clip]{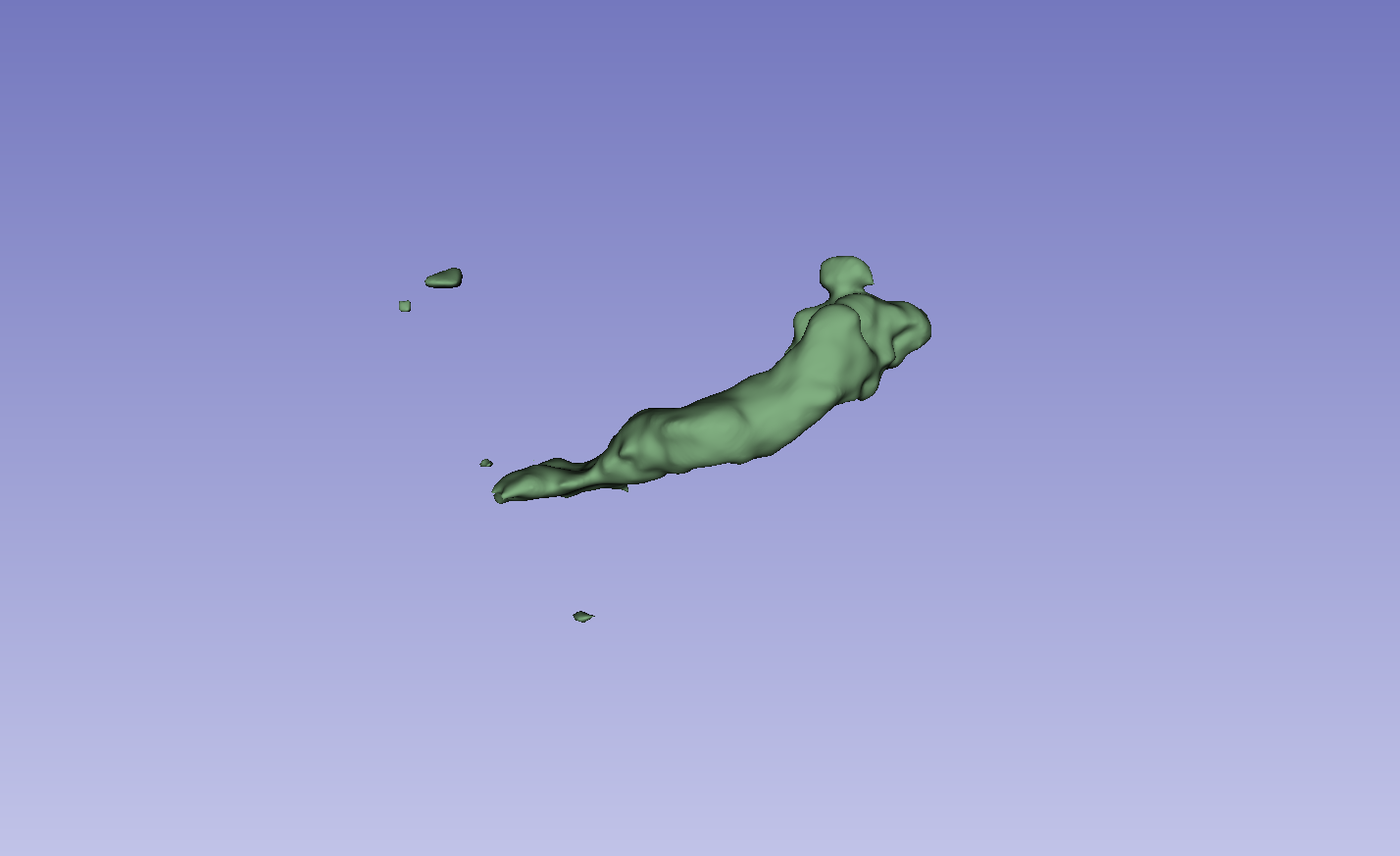} &
    \hspace{-1em} 
    \includegraphics[width=0.25\textwidth, trim=219.5pt 140pt 219.5pt 110pt, clip]{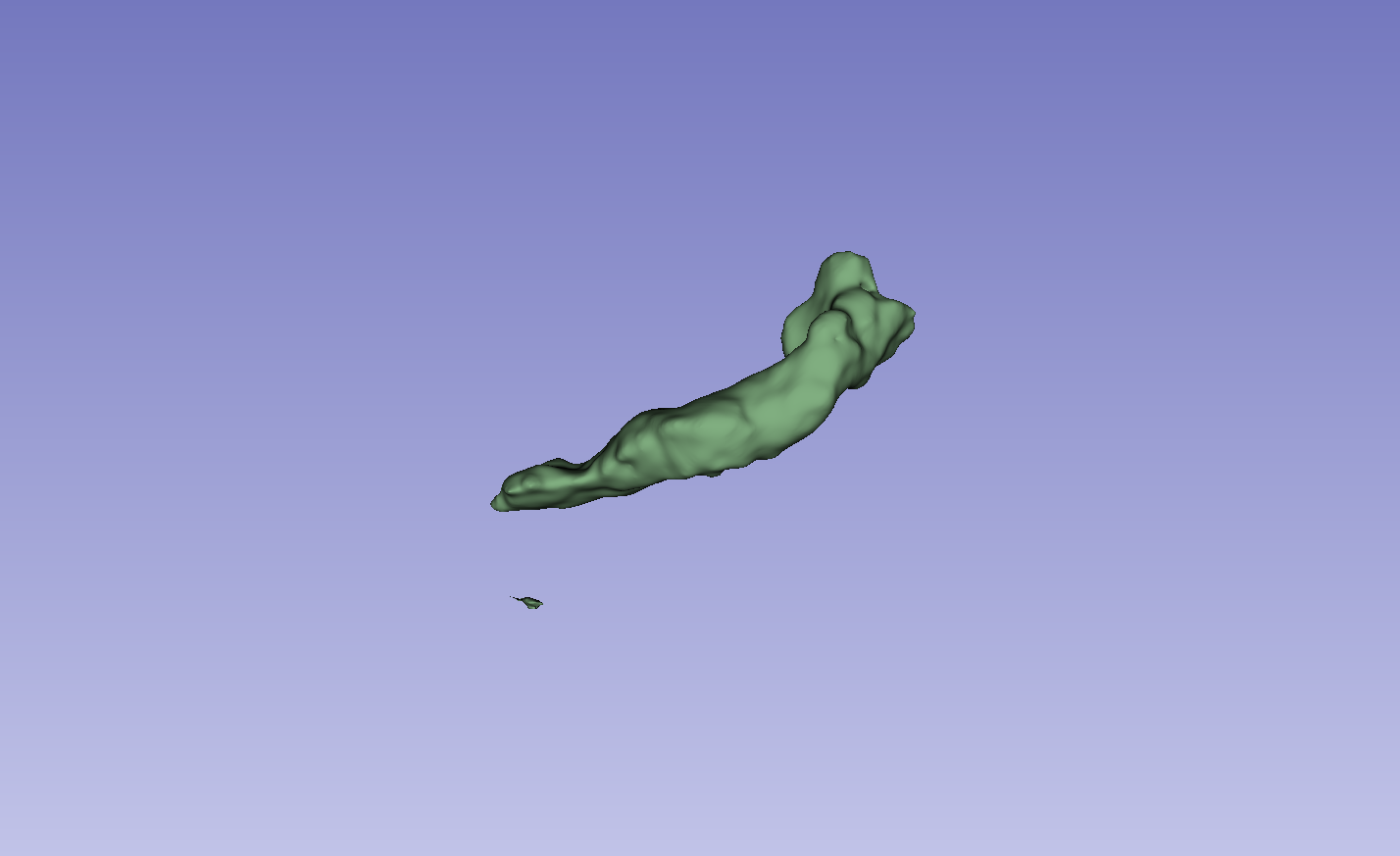} &
    \hspace{-1em} 
    \includegraphics[width=0.25\textwidth, trim=219.5pt 140pt 219.5pt 110pt, clip]{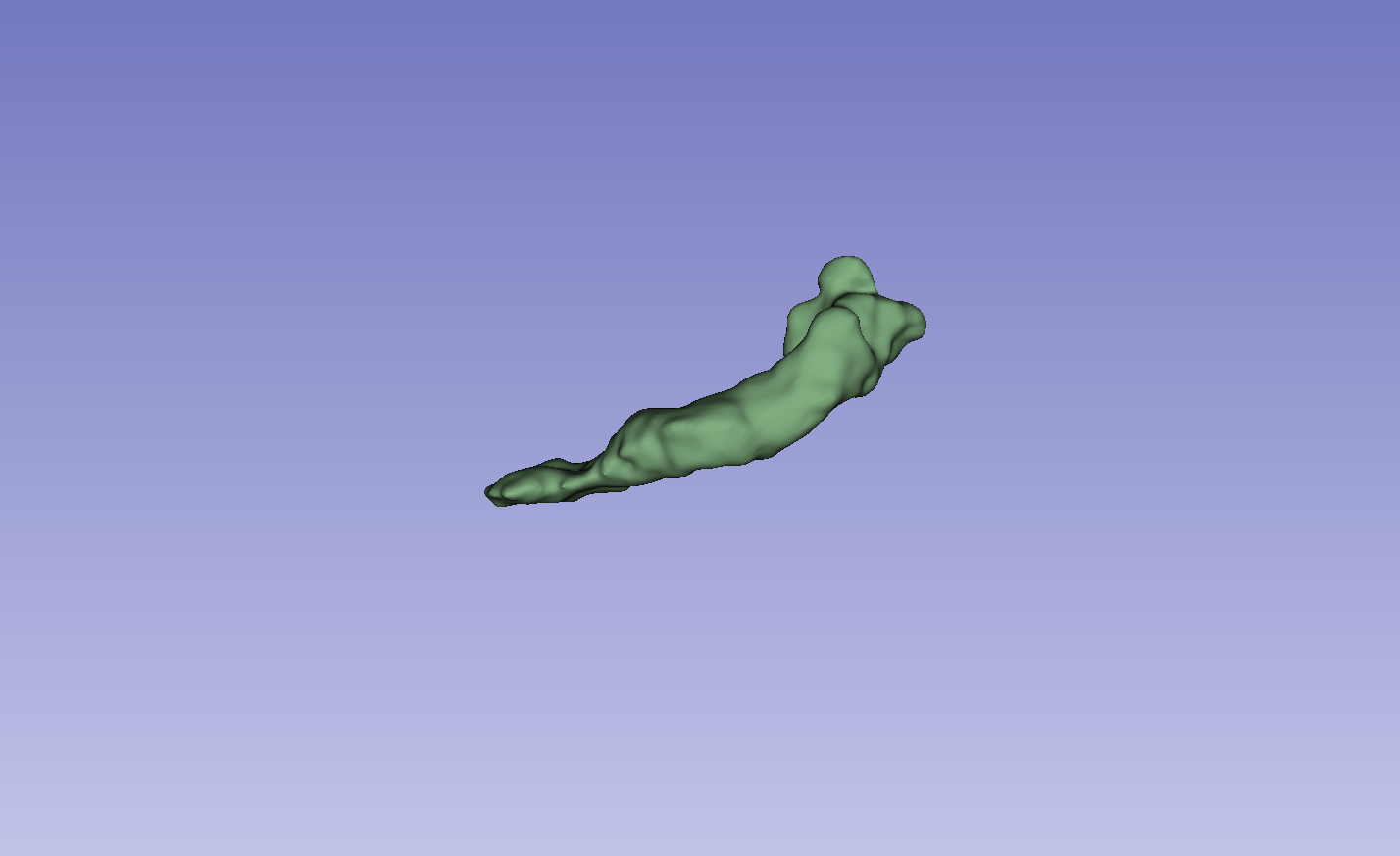} \\
    {\small (a) Ground Truth} & {\small(b) MC-Net} & {\small(c) MCF} & {\small(e) Proposed Method} 
    \end{tabular}
    }
\caption{Visual comparison of segmentation results: the first and the second rows show the left atrium (LA) and pancreas, respectively.} \label{fig:visualresults}
\end{figure}
\vspace{-0.9em}
The comparison of our proposed method with SOTA techniques on the left atrial dataset is provided in Table~\ref{table1}. Our method shows significant improvements in all metrics, with a substantial enhancement in organ voxel detection, specifically DSC and Jacard. Compared to MCF, our proposed method exhibits noteworthy enhancements, with an increase in DSC from 88.71 to 89.10 and Jaccard index from 80.41 to 81.62. Furthermore, our approach maintains low-performance variance, contributing to its stability and reliability. \Cref{fig:visualresults} presents the visual results of our proposed method compared to other methods for left atrial segmentation. These visual results showcase higher overlap with ground truth labels and fewer false segmentations, highlighting the finer details captured by our approach.
Our method showcases strong performance on the Pancreas dataset as well, as presented in \Cref{table2}.
\Cref{fig:visualresults} provides additional insights into the segmentation results, underscoring the impact of the suggested modules for enhancing the overall segmentation quality. In greater detail, our approach generates sharper edges and more precise boundary separation then the MCF and MC-Net methods. This highlights its effectiveness in improving the reliability of object boundary predictions and distinguishing the organ of interest from the background.

We also performed an ablation study on the LA dataset to thoroughly assess the impact of the regularization and contrastive loss components of our method. Notably, removing the regularization module resulted in a substantial decrease of 0.5 in the DSC. Similarly, removing the contrastive loss resulted in a more significant drop of 0.73 in the DSC score. These findings underscore the critical roles played by both the regularization and contrastive loss components in enhancing the performance of our method.

\begin{table}[!htb]
    \centering
    \caption{Comparison of results using the Pancreas dataset (CT).}
    \label{table2}
    \begin{threeparttable}
    \resizebox{\columnwidth}{!}{
    \begin{tabular}{c|cccc}
        \hline 
        \hline
        Method & Dice(\%)$\uparrow$ & Jaccard(\%)$\uparrow$ & 95HD(voxel)$\downarrow$ & ASD(voxel)$\downarrow$ \\
        \hline
        MT \cite{tarvainen2017mean} & 74.43 $\pm$ 0.024 & 60.53 $\pm$ 0.030 & 14.93 $\pm$ 2.000 & 4.61 $\pm$ 0.929 \\
        UA-MT \cite{yu2019uncertainty} & 74.01 $\pm$ 0.029 & 60.00 $\pm$ 3.031 & 17.00 $\pm$ 3.031 & 5.19 $\pm$ 1.267 \\
        SASSNet \cite{li2020shape} & 73.57 $\pm$ 0.017 & 59.71 $\pm$ 0.020 & 13.87 $\pm$ 1.079 & 3.53 $\pm$ 1.416 \\
        DTC \cite{luo2021semi} & 73.23 $\pm$ 0.024 & 59.18 $\pm$ 0.027 & 13.20 $\pm$ 2.241 & 3.81 $\pm$ 0.953 \\
        MC-Net \cite{wu2021semi} & 73.73 $\pm$ 0.019 & 59.19 $\pm$ 0.021 & 13.65 $\pm$ 3.902 & 3.92 $\pm$ 1.055 \\
        MCF \cite{wang2023mcf} & 75.00 $\pm$ 0.026 & 61.27 $\pm$ 0.030 & 11.59 $\pm$ 1.611 & 3.27 $\pm$ 0.919 \\
        \rowcolor[HTML]{C8FFFD}
        Our Method & 76.40 $\pm$ 0.018 & 62.96 $\pm$ 0.027 & 10.69 $\pm$ 1.603 & 2.79 $\pm$ 0.0954 \\
        \hline
        \hline
    \end{tabular}}
    \end{threeparttable}
\end{table}
\vspace{-1em}

\section{Conclusion}
This paper presents a novel dual-stream network for semi-supervised semantic segmentation that leverages labeled and unlabeled imaging data. Our approach focuses on reducing the problem of unreliable predictions by integrating contrastive learning and error correction mechanisms. We outperformed SOTA techniques on CT and MRI images.


\bibliographystyle{IEEEbib}
\bibliography{refs.bib}

\begin{thebibliography}{10}

\bibitem{azad2022medical}
Reza Azad, Ehsan~Khodapanah Aghdam, Amelie Rauland, Yiwei Jia, Atlas~Haddadi Avval, Afshin Bozorgpour, Sanaz Karimijafarbigloo, Joseph~Paul Cohen, Ehsan Adeli, and Dorit Merhof,
\newblock ``Medical image segmentation review: The success of u-net,''
\newblock {\em arXiv preprint arXiv:2211.14830}, 2022.

\bibitem{azad2023foundational}
Bobby Azad, Reza Azad, Sania Eskandari, Afshin Bozorgpour, Amirhossein Kazerouni, Islem Rekik, and Dorit Merhof,
\newblock ``Foundational models in medical imaging: A comprehensive survey and future vision,''
\newblock {\em arXiv preprint arXiv:2310.18689}, 2023.

\bibitem{antonelli2022medical}
Michela Antonelli, Annika Reinke, Spyridon Bakas, Keyvan Farahani, Annette Kopp-Schneider, Bennett~A Landman, Geert Litjens, Bjoern Menze, Olaf Ronneberger, Ronald~M Summers, et~al.,
\newblock ``The medical segmentation decathlon,''
\newblock {\em Nature communications}, vol. 13, no. 1, pp. 1--13, 2022.

\bibitem{srivastava2022efficient}
Abhishek Srivastava, Debesh Jha, Elif Keles, Bulent Aydogan, Mohamed Abazeed, and Ulas Bagci,
\newblock ``An efficient multi-scale fusion network for 3d organ at risk (oar) segmentation,''
\newblock {\em arXiv preprint arXiv:2208.07417}, 2022.

\bibitem{cps}
Xiaokang Chen, Yuhui Yuan, Gang Zeng, and Jingdong Wang,
\newblock ``Semi-supervised semantic segmentation with cross pseudo supervision,''
\newblock in {\em IEEE Conference on Computer Vision and Pattern Recognition}, 2021.

\bibitem{tarvainen2017mean}
Antti Tarvainen and Harri Valpola,
\newblock ``Mean teachers are better role models: Weight-averaged consistency targets improve semi-supervised deep learning results,''
\newblock in {\em Advances in Neural Information Processing Systems}, 2017.

\bibitem{wang2023mcf}
Yongchao Wang, Bin Xiao, Xiuli Bi, Weisheng Li, and Xinbo Gao,
\newblock ``Mcf: Mutual correction framework for semi-supervised medical image segmentation,''
\newblock in {\em Proceedings of the IEEE/CVF Conference on Computer Vision and Pattern Recognition}, 2023, pp. 15651--15660.

\bibitem{cct}
Yassine Ouali, C{\'e}line Hudelot, and Myriam Tami,
\newblock ``Semi-supervised semantic segmentation with cross-consistency training,''
\newblock in {\em IEEE Conference on Computer Vision and Pattern Recognition}, 2020.

\bibitem{lee2013pseudo}
Dong-Hyun Lee et~al.,
\newblock ``Pseudo-label: The simple and efficient semi-supervised learning method for deep neural networks,''
\newblock in {\em International Conference on Machine Learning Workshop}, 2013, vol.~3, p. 896.

\bibitem{basak2023pseudo}
Hritam Basak and Zhaozheng Yin,
\newblock ``Pseudo-label guided contrastive learning for semi-supervised medical image segmentation,''
\newblock in {\em Proceedings of the IEEE/CVF Conference on Computer Vision and Pattern Recognition}, 2023, pp. 19786--19797.

\bibitem{chaitanya2023local}
Krishna Chaitanya, Ertunc Erdil, Neerav Karani, and Ender Konukoglu,
\newblock ``Local contrastive loss with pseudo-label based self-training for semi-supervised medical image segmentation,''
\newblock {\em Medical Image Analysis}, vol. 87, pp. 102792, 2023.

\bibitem{bai2023bidirectional}
Yunhao Bai, Duowen Chen, Qingli Li, Wei Shen, and Yan Wang,
\newblock ``Bidirectional copy-paste for semi-supervised medical image segmentation,''
\newblock in {\em Proceedings of the IEEE/CVF Conference on Computer Vision and Pattern Recognition}, 2023, pp. 11514--11524.

\bibitem{luo2021semi}
Xiangde Luo, Jieneng Chen, Tao Song, and Guotai Wang,
\newblock ``Semi-supervised medical image segmentation through dual-task consistency,''
\newblock in {\em Proceedings of the AAAI conference on artificial intelligence}, 2021, vol.~35, pp. 8801--8809.

\bibitem{wu2021semi}
Yicheng Wu, Minfeng Xu, Zongyuan Ge, Jianfei Cai, and Lei Zhang,
\newblock ``Semi-supervised left atrium segmentation with mutual consistency training,''
\newblock in {\em Medical Image Computing and Computer Assisted Intervention--MICCAI 2021: 24th International Conference, Strasbourg, France, September 27--October 1, 2021, Proceedings, Part II 24}. Springer, 2021, pp. 297--306.

\bibitem{zhang2021flexmatch}
Bowen Zhang, Yidong Wang, Wenxin Hou, Hao Wu, Jindong Wang, Manabu Okumura, and Takahiro Shinozaki,
\newblock ``Flexmatch: Boosting semi-supervised learning with curriculum pseudo labeling,''
\newblock in {\em Advances in Neural Information Processing Systems}, 2021, vol.~34.

\bibitem{guo2017calibration}
Chuan Guo, Geoff Pleiss, Yu~Sun, and Kilian~Q Weinberger,
\newblock ``On calibration of modern neural networks,''
\newblock in {\em International Conference on Machine Learning}. PMLR, 2017, pp. 1321--1330.

\bibitem{xiong2021global}
Zhaohan Xiong, Qing Xia, Zhiqiang Hu, Ning Huang, Cheng Bian, Yefeng Zheng, Sulaiman Vesal, Nishant Ravikumar, Andreas Maier, Xin Yang, et~al.,
\newblock ``A global benchmark of algorithms for segmenting the left atrium from late gadolinium-enhanced cardiac magnetic resonance imaging,''
\newblock {\em Medical image analysis}, vol. 67, pp. 101832, 2021.

\bibitem{roth2015deeporgan}
Holger~R Roth, Le~Lu, Amal Farag, Hoo-Chang Shin, Jiamin Liu, Evrim~B Turkbey, and Ronald~M Summers,
\newblock ``Deeporgan: Multi-level deep convolutional networks for automated pancreas segmentation,''
\newblock in {\em MICCAI 2015}. Springer, 2015, pp. 556--564.

\bibitem{yu2019uncertainty}
Lequan Yu, Shujun Wang, Xiaomeng Li, Chi-Wing Fu, and Pheng-Ann Heng,
\newblock ``Uncertainty-aware self-ensembling model for semi-supervised 3d left atrium segmentation,''
\newblock in {\em MICCAI 2019}. Springer, 2019, pp. 605--613.

\bibitem{li2020shape}
Shuailin Li, Chuyu Zhang, and Xuming He,
\newblock ``Shape-aware semi-supervised 3d semantic segmentation for medical images,''
\newblock in {\em MICCAI 2020}. Springer, 2020.

\end{thebibliography}

\end{document}